# Transformer-based Multi-task Learning for Disaster Tweet Categorisation

**Congcong Wang***
School of Computer Science
University College Dublin
congcong.wang@ucdconnect.ie

**Paul Nulty**
School of Computer Science
University College Dublin
paul.nulty@ucd.ie

**David Lillis**
School of Computer Science
University College Dublin
david.lillis@ucd.ie

**ABSTRACT**

Social media has enabled people to circulate information in a timely fashion, thus motivating people to post messages seeking help during crisis situations. These messages can contribute to the situational awareness of emergency responders, who have a need for them to be categorised according to information types (i.e. the type of aid services the messages are requesting). We introduce a transformer-based multi-task learning (MTL) technique for classifying information types and estimating the priority of these messages. We evaluate the effectiveness of our approach with a variety of metrics by submitting runs to the TREC Incident Streams (IS) track: a research initiative specifically designed for disaster tweet classification and prioritisation. The results demonstrate that our approach achieves competitive performance in most metrics as compared to other participating runs. Subsequently, we find that an ensemble approach combining disparate transformer encoders within our approach helps to improve the overall effectiveness to a significant extent, achieving state-of-the-art performance in almost every metric. We make the code publicly available so that our work can be reproduced and used as a baseline for the community for future work in this domain[1].

**INTRODUCTION**

According to Guha-Sapir et al. (2012), an average of 50,000 people worldwide die from natural disasters annually. Effective and accurate knowledge of how an incident is unfolding (known as "situational awareness" (SA)) helps response services to take timely preventative measures to remedy a crisis situation (Endsley 2017). Social media can provide real-time contact and communication between emergency aid centres and those in the vicinity of incidents, and as such has been identified as an important tool in establishing SA (S. Vieweg et al. 2010; S. E. Vieweg 2012).

Substantial work has been undertaken to examine the possibility of SA on social media (Lambert et al. 2005; Norris 2006; Palen and S. B. Liu 2007). According to one study, 69% of people believe that emergency response operators should monitor their sites and social media accounts, and respond promptly during a crisis[2]. A recent study also shows that around 10% of emergency-related posts on Twitter[3] are actionable and around 1% are critical (McCreadie et al. 2019). Manual identification of useful messages is unfeasible, particularly as the quantity of messages tends to increase explosively during a crisis (Lambert et al. 2005; Palen and S. B. Liu 2007). Thus computational linguistics techniques have been sought for automating the classification of such messages (Imran et al. 2013; Olteanu, Castillo, et al. 2014; Olteanu, S. Vieweg, et al. 2015; Zahra et al. 2020).

---

*corresponding author
[1] https://github.com/wangcongcong123/crisis-mtl
[2] https://mashable.com/2011/02/11/social-media-in-emergencies
[3] https://twitter.com

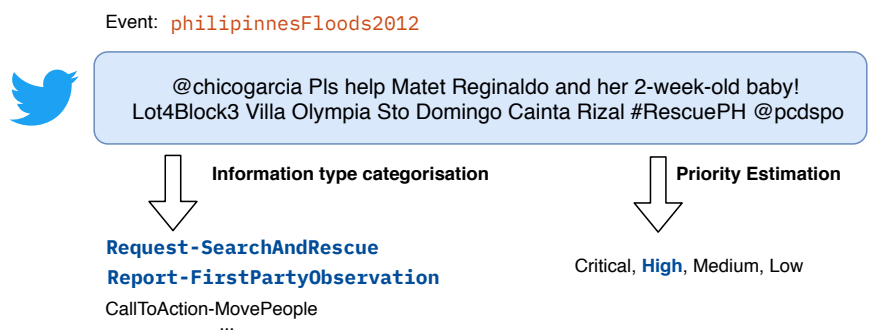

**Figure 1. An example of disaster tweet categorisation**

The TREC Incident Streams (IS) track (McCreadie et al. 2019; McCreadie et al. 2020) is an initiative designed for the categorisation of crisis-related tweets. Figure 1 presents an example of disaster tweet categorisation, illustrating two sub-tasks of the IS track. Given a crisis-related tweet, participants are asked to both ascribe information types and estimate its priority. Human assessors are employed to annotate such crisis-related tweets, forming a crisis dataset known as TREC-IS dataset, which has been growing since 2018. The tweets are labeled with one or more information types (i.e. what information needs a user-posted tweet is about that can be useful for emergency responders in making aid-relevant decisions). TREC-IS defines an ontology of 25 information types, covering major aspects of such information needs, among which 6 are defined as "actionable" and the rest are "non-actionable". These are detailed in Table 1. In addition to the information types, each tweet is also assigned a level of priority, indicating the criticality of the tweet. The priority can be "critical", "high", "medium" or "low". The TREC-IS dataset provides a benchmark dataset for studies on crisis-related messages processing or categorisation, which is crucial to the community leveraging social media for emergency response. The IS track also proposes a research-friendly standardised evaluation methodology for systematically measuring the performance of participating systems. The official website[4] contains all details relevant to the IS track.

This paper presents our method in both the information type categorisation task and the priority estimation task. We describe and analyse the work that we submitted as part of our participation in the IS track, and also additional research work that has continued beyond this. In recent times, transformer-based deep networks (Vaswani et al. 2017) such as BERT (Devlin et al. 2019) and ELECTRA (Clark et al. 2020) have become widely adopted due to their strong performance in short message processing. We propose a transformer-based multi-task learning (MTL) technique that categorises the crisis tweets through a joint learning of two sub-tasks, namely, information type classification and priority estimation. The contributions of our work are multifaceted and summarised as follows.

- We propose a transformer-base MTL approach for crisis tweet categorisation. The experimental results show it outperforms both transformer-based single-task learning and traditional machine learning baselines.

- We introduce a simple ensemble technique that leverages the joint predictions of multiple MTL models for crisis tweet categorisation, which is demonstrated to be very effective as compared to the individual MTL models.

- We submit runs based on our transformer-based MTL approach to the IS-track. The returned results present that our MTL runs overall outperform the competitive participating runs. Subsequently, our ensemble runs notably outperform the participating runs in almost every aspect of performance.

## RELATED WORK

A number of existing computational methods have been applied to the problem of automatic classification of crisis-related social media messages. These methods can be broadly divided into three categories. The classic methods fall into traditional machine learning (ML) algorithms, which are known for efficient computation and explainable feature-based predictions. For example, Caragea, McNeese, et al. (2011) employed the classic SVM algorithm for classifying messages from the 2010 Haiti Earthquake according to some important information types including people trapped, food shortage, etc. Li et al. (2018) applied Logistic Regression and Naïve Bayes for crisis tweet classification with an adaptation approach.

---

[4]http://dcs.gla.ac.uk/~richardm/TREC_IS/

|  | Category | Description |
| --- | --- | --- |
| Actionable | Request-GoodsServices | The user is asking for a particular service or physical good |
|  | Request-SearchAndRescue | The user is requesting a rescue (for themselves or others) |
|  | Report-NewSubEvent | The user is reporting a new occurence that public safety officers need to respond to |
|  | Report-ServiceAvailable | The user is reporting that they or someone else is providing a service |
|  | CallToAction-MovePeople | The user is asking people to leave an area or go to another area |
|  | Report-EmergingThreats | The user is reporting a potential problem that may cause future loss of life or damage |
| Non-actionable | CallToAction-Volunteer | The user is asking people to volunteer to help the response effort |
|  | CallToAction-Donations | The user is asking people to donate goods/money |
|  | Report-Weather | The user is providing a weather report (current or forcast) |
|  | Report-Location | The post contains information about the user or observation location |
|  | Request-InformationWanted | The user is requesting information |
|  | Report-FirstPartyObservation | The user is giving an eye-witness account |
|  | Report-ThirdPartyObservation | The user is reporting a information that they recieved from someone else |
|  | Report-MultimediaShare | The user is sharing images or video |
|  | Report-Factoid | The user is relating some facts, typically numerical |
|  | Report-Official | An official report by a government or public safety representative |
|  | Report-News | The post is a news report providing/linking to current/continious coverage of the event |
|  | Report-CleanUp | A report of the clean up after the event |
|  | Report-Hashtags | Reporting which hashtags correspond to each event |
|  | Report-OriginalEvent | A report of the original event occuring. |
|  | Other-ContextualInformation | The post contains contextual information that can help understand the event, but is not about the event itself |
|  | Other-Advice | The author is providing some advice to the public |
|  | Other-Sentiment | The post is expressing some sentiment about the event |
|  | Other-Discussion | Users are discussing the event |
|  | Other-Irrelevant | The post is unrelated to the event or contains no information |

**Table 1. Actionable and non-actionable information types in the TREC-IS dataset (McCreadie et al. 2019).**

Deep learning methods based on neural networks have been applied to good effect to various problems related to short message processing. In Neppalli et al. (2018), traditional ML methods (Naïve Bayes with hand-crafted features) were compared to deep methods (CNN and RNN), finding that deep methods outperform the classic ones (particularly CNN). Several other studies applied CNNs for categorising crisis-related data on social networks (Nguyen et al. 2017; Caragea, Silvescu, et al. 2016). Kumar et al. (2020) proposed a multi-modal approach for identifying informative crisis-related tweets using both text and image data, which are trained by a long short-term memory (LSTM) and VGG-16 network respectively. Another important study of this kind is that of Chowdhury et al. (2020), who proposed a LSTM-based multi-task model for identifying hashtags in crisis tweets.

Transformers are a series of attention-based neural network models (Wolf et al. 2019), originating from the original transformer paper (Vaswani et al. 2017). In the field of transfer learning for NLP, fine-tuning the transformer-based models such as BERT (Devlin et al. 2019), ELECTRA (Clark et al. 2020), DistilBERT (Sanh et al. 2019), etc., has become the mainstream way to achieve state-of-the-art performance for various language tasks in different domains. For example, in the domain of crisis message processing, J. Liu et al. (2020) proposed CrisisBERT, which fine-tuned pre-trained BERT to achieve state-of-the-art performance in crisis events identification. Chowdhury et al. (2020) applied BERT for multi-label crisis-related tweets classification with Manifold Mixup in cross-linguistic settings. Since MTL has achieved remarkable success in NLP applications (Zhang and Yang 2017), recent years have seen many work into applying transformers via MTL for text classification tasks in similar domains. For example, Xue et al. (2019) applied BERT for joint learning on name entity recognition (NER) and relation extraction classification tasks from a Chinese medical text corpus. Similarly in the domain of medical text mining, instead of fine-tuning jointly on two tasks from a single corpus, Peng et al. (2020) studied the effectiveness of BERT jointly fine-tuned on multiple corpora from different types of tasks including text similarity, NER, text inference, etc.

Inspired by the work using MTL with BERT in similar domains, we extend BERT to multiple transformers and fine-tune them for crisis tweet categorisation in a MTL way. In addition, we investigate the power of an ensemble of multiple transformer-based MTL models, which has not been seen in the literature to the best of our knowledge.

## METHOD

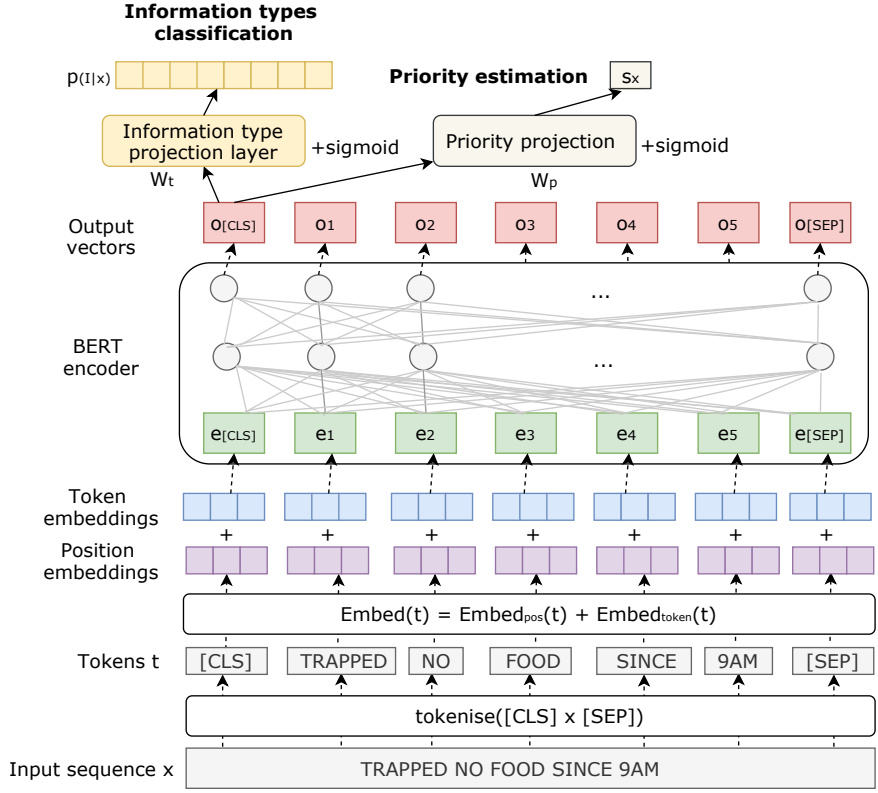

**Figure 2. The overview architecture of our transformer-based multi-task learning approach.**

Based on the objectives of disaster tweet categorisation as described, we break them down to two downstream sub-tasks by fine-tuning the transformer-encoder models[5] in the manner of multi-task learning (MTL). Figure 2 depicts the overview architecture of the MTL approach for the information type categorisation task (denoted by *ITC*) and the priority estimation task (denoted by *PE*).

The objective of *ITC* is to predict the probability: $p_{(I_j|x)}$ referring to the likelihood of an input message $x$ being assigned to the information type $I_j$. Since $x$ can be assigned to one or more information types, it is taken as a multi-label classification problem, estimated by the following equation.

$$p_{(I|x)} = \sigma(f(x)) \qquad (1)$$

$$\sigma(a) = \frac{1}{(1 + e^{-a})} \qquad (2)$$

Where $p_{(I|x)}$ is the estimated probability distribution across all information types $I : \{I_o, I_1, ...I_j, ..., I_n\}$ and $n$ is the number of information types. In our method, an information type $I_j$ is assigned to $x$ when $p_{(I_j|x)} > 0.5$.

To learn the function $f(\cdot)$, we add an information type projection layer. We choose BERT's `[CLS]` token output vector $o_{[CLS]}$, denoted by $\text{BERT}_{CLS}(x)$, as the input of the projection layer. The projection process is formulated as follows.

$$f(x) = \text{BERT}_{CLS}(x) W_t \qquad (3)$$

Where $W_t \in \mathbb{R}^{d_{\text{model}} \times n}$ is the learnable parameters of the projection linear layer and $d_{\text{model}}$ is the hidden state dimension of the transformer encoder (i.e., the BERT encoder in Figure 2). Considering the BERT-style transformer as an upstream encoder, this outputs a representation vector for each token of the input sequence using the multi-head self-attention mechanism described in Vaswani et al. (2017). The lower part of Figure 2 illustrates this process and it can also be expressed mathematically as

$$\text{BERT}_t(x) = \text{Encoder}_{up}(\text{Embed}(t)))  \quad (4)$$

where $t : \{\text{CLS}, t_1, t_2.., \text{SEP}\}$ refers to the word pieces tokenised from the input sequence $x$. These are represented by combining two types of vectors, namely position embeddings and token embeddings[6], before being fed to the self-attention-based BERT encoder. It can be formulated as follows.

---

[5]Here we take BERT as an example, although the approach generalises to other transformer encoders as discussed below.

$$\text{Embed}(t) = \text{Embed}_{pos}(t) + \text{Embed}_{token}(t) \quad (5)$$

$$t = \text{tokenize}(\text{[CLS]} \ x \ \text{[SEP]}) \quad (6)$$

In a similar way to *ITC*, the *PE* task is treated as a regression task (due to the dependency between priority levels). In *PS*, a numeric score $s \in [0, 1]$ is assigned to the input example $x$ for quantifying its priority level, which is defined as follows.

$$s_x = \sigma(g(x)) \quad (7)$$

To learn the function $g(\cdot)$, a priority projection layer is added, which transforms the [CLS] token output vector to the priority score, formulated as follows.

$$g(x) = \text{BERT}_{\text{CLS}}(x) W_p \quad (8)$$

Where $W_p \in \mathbb{R}^{d_{\text{model}} \times 1}$ is the learnable parameters of the priority projection layer and $d_{\text{model}}$ is the hidden state dimension of the BERT-style transformer encoder. Based on this equation, we can see the two tasks share the upstream encoder, i.e., mathematically, using the same layers of millions of parameters for encoding input tokens $t : \{\text{CLS}, t_1, t_2.., \text{SEP}\}$ to output vectors $o : \{o_{\text{[CLS]}}, o_1, o_2.., o_{\text{[SEP]}}\}$. The motivation behind this is two-fold. Firstly, parameter sharing between multiple tasks is likely to enable one task to share its learnt knowledge with another Zhang and Yang 2017. Secondly, it is faster to make predictions at inference time than training two separate models for every task.

To update the parameters of the projection layers as well as fine-tune these from the transformer-encoder, we define the combination loss function as follows.

$$L_{\text{total}} = \lambda L_{\text{it}} + (1 - \lambda) L_{\text{pri}} \quad (9)$$

Where $\lambda$ is the loss parameter for $L_{\text{it}}$ and $L_{\text{pri}}$ adjusting their weight, and $L_{\text{it}}$ is the loss between predicted information types and human-annotated labels. This tells that it will be the priority estimation single task and information type classification task fine-tuning when the $\lambda$ is set to be 0 and 1 respectively.

The following formulates the loss computation for one training example. For a mini-batch of examples, the final loss is averaged over the examples.

$$L_{\text{it}} = \sum_{I_j \in I} -b(I_j) \log(p(I_j|x)) - (1 - b(I_j)) \log(1 - p(I_j|x)) \quad (10)$$

Where $b(I_j)$ is 0 or 1, indicating if the information type $I_j$ is assigned to the example $x$ (information type ground truth) and $p(I_j|x)$ is the probability score for predicting $x$ to be $I_j$.

The $L_{\text{pri}}$ in Equation 9 refers to the priority loss calculated using Mean Squared Error, which is formally defined as follows.

$$L_{\text{pri}} = (m(r) - s_x)^2 \quad (11)$$

Where $r$ is the priority level assigned to $x$ (priority ground truth) and $m(\cdot)$ is the mapping function converting categorical priority levels to numeric scores according to the schema: $\{Critical : 1.0, High : 0.75, Medium : 0.5, Low : 0.25\}$. At prediction time, the mapping function is used in reverse to convert the predicted numeric score to a categorical priority level. For example, the priority of $x$ is predicted to be critical if $s_x$ lies between 1 and 0.75.

---

[6]BERT also features segment embedding, which is omitted here since only a single sequence is used as input in our problem (i.e. segment IDs are zeros).

## EXPERIMENT

This section presents the experimental components that are necessary for results demonstration and discussion, including dataset construction, training details, and evaluation metrics.

### Datasets

Having run several editions since 2018, the TREC-IS dataset has grown to contain approximately 40,000 examples, which are divided into four subsets based on the four editions of the track that have run to date. The subsets are **2018**, **2019A**, **2019B**, and **2020A**, with each consisting of tweets across different crisis events and the events between the subsets are not overlapping. Table 2 presents the main characteristics of these subsets. As this illustrates, the examples are short, with only around 0.1% more than 128 BERT-based word pieces. To fit our experimental purposes, we use the 2020A subset set as the test set (6,658 examples) that consists of tweets from 15 crisis events occurred in 2020. For the training dataset, we first generate a combined set containing all examples from 2018, 2019A and 2019B. We then sample 10% from the combined set to form the development set for model selection and the remaining 90% to form the training set. This results in 30,420 training examples and 3,381 for model selection.

| Subsets | Size | No. events | Avg. length | Beyond 128 |
|---|---|---|---|---|
| 2018 | 17581 | 15 | 35 | 0 |
| 2019A | 7098 | 6 | 37 | 0 |
| 2019B | 9122 | 6 | 48 | 0.09% |
| 2020A | 6658 | 15 | 42 | 0.60% |
| Total | 40459 | 42 | 40 | 0.10% |

**Table 2. Statistics of TREC-IS sub-datasets.** *No. events* means the number of crises for a set. *Avg. length* refers to the average number of tokens of each example and *Beyond 128* indicates the percentage of examples with more than 128 tokens, counted as BERT-style word pieces.

### Training

Training mainly involves fine-tuning the pre-trained transformer encoders to the joint downstream task of information type multi-label classification and priority estimation. For hyper-parameter selection, we experimented with a grid search over learning rate: $lr \in \{5e-4, 2e-4, 1e-4, 5e-5, 2e-5, 1e-5\}$ and mini-batch size: $bs \in \{8, 16, 32, 64\}$. Finally, we chose $lr$ and $bs$ to be $5e-5$ and 32 respectively as they perform better empirically.

Following the work in a similar domain (J. Liu et al. 2020), we use the Adam optimiser (Kingma and Ba 2015) to update model parameters and a linear scheduler for dynamically updating learning rate, with 10% warm-up ratio of the total training steps[7]. In addition, we set $\lambda$ from Equation 9 to be 0.5, giving equal weight to the information type loss and priority loss during fine-tuning [8]. We set the maximum sequence length to be 128, since a negligible number of training examples have more than 128 tokens, as presented in Table 2. All the runs in our experiment are accelerated by a `RTX-2080Ti` and a `RTX-2070 super` GPU.

### Evaluation Metrics

The IS track asks participants to submit their systems' runs on test tweets for which information types and priority levels are later judged by human assessors. Moreover, the runs rank the submitted tweets per event type according to estimated priority scores so that they can be evaluated uniformly. We evaluate our system's effectiveness using the same evaluation metrics as are used in the IS track (McCreadie et al. 2020). The metrics are broadly divided into four categories evaluating different aspects of a system's effectiveness: **Ranking (NDCG), Alerting Worth (AW-HC, AW-A), Information Feed Categorisation (CF1-H, CF1-A, cacc) Prioritisation (PF1-H, PF1-A)**. Each metric in these categories is represented by a numeric score within a certain range where higher is better, which are briefly described below.

- **Ranking** (range 0 to 1): In this category, **NDCG** (Järvelin and Kekäläinen 2002) is the priority-centric metric used to evaluate the quality of submitted test tweets ranked by priority scores. By default, it measures the top 100 submitted tweets per event. A high NDCG score implies that the system has achieved a good quality of priority-based ranking when compared with human assessors.

---

[7] We fine-tuned 12 epochs totalling approximately 3.4k steps evaluated every 400 steps as we observed no further performance improvement when increasing the epoch number.
[8] This is based on our pilot study that found $\lambda$ being 0.5 gives overall good performance in both tasks

- **Alerting Worth** (range −1 to 1): This is inspired by the alerting use case in a real-world emergency response system. It not only measures the effectiveness of a system in generating true alerts but also penalises the system in generating consecutive false alerts that would make end users lose trust in the system. Two components of Alerting Worth are **AW-HC** and **AW-A**, which measure the effectiveness of true alerts within the scope of tweets judged to be critical or high, and within the scope of all priority-level tweets respectively.

- **Information Feed Categorisation** (range 0 to 1): This is used to evaluate the aspect of information type classification performance by a system. To better reflect a system's utility to emergency response officers, it consists of three specific metrics, Actionable F1, All F1, and Accuracy, denoted by **CF1-H**, **CF1-A**, and **Cacc** respectively. CF1-A macro-averages the F1 scores across all information types, while CF1-H macro-averages the F1 scores only across the 6 actionable types as presented in Table 1. The two metrics indicate the performance of information type categorisation by only taking the target class per information type into account. Cacc computes the categorisation accuracy micro-averaged across information types, which offers a general view of categorisation performance.

- **Prioritisation** (range 0 to 1): This is applied to measure the performance of priority level predictions, consisting of two specific metrics, Actionable F1 and All F1, abbreviated to **PErr-H** and **PErr-A**. Both are computed by averaging the macro-F1 scores on priority label predictions per information type. Unlike PErr-A, which averages the F1 scores for all information types, PErr-H averages the F1 scores for actionable types only.

As a general reference, the metrics in the category of information feed categorisation indicate the effectiveness of the information type classification task while the rest reveal the effectiveness of the priority estimation task. In addition to the above mentioned official metrics, we add a harmonic mean metric (**HarM**) to the list as an indicator of overall performance across all metrics. In calculating this harmonic mean, AW-HC and AW-A are first normalised to lie between 0 and 1.

## RESULTS AND DISCUSSION

The major objective of our study is to answer essential questions that we consider to be beneficial for further work into this problem domain by the community. We report and discuss our experimental results to answer the following research questions in the three sub-sections that follow.

- **RQ1:** Does multi-task learning add benefits compared to single-task learning baselines?
- **RQ2:** How do the pre-trained transformer encoders influence performance? Does ensemble learning help?
- **RQ3:** How does our approach perform compared to top runs submitted to the IS track?

### RQ1: Single tasks learning baselines

To answer the first question, we use BERT_base[9] as a baseline to fine-tune two models independently for the *ITC* and *PE* tasks in a single task learning (STL) scenario. The loss functions of Equation 10 and Equation 11 are used separately to train the two single task models. Our MTL scenario uses the joint loss function of Equation 9. In addition to the transformer-based deep baseline we also consider traditional machine learning (ML) algorithms such as Logistic Regression (LR) to be strong baselines in this problem domain, as evidenced in previous work (McCreadie et al. 2020; C. Wang and Lillis 2020). Hence, we also train two separate LR-based classifiers for the two single tasks. To make this baseline as strong as possible, we use the development set of TS to test different options in configuring LR, including $C \in \{0.01, 0.1, 1, 1.0, 10, 100\}$, $ngram\_range \in \{(1), (1, 2), (1, 3)\}$ and $weighting \in \{\text{count}, \text{TFIDF}\}$. Following empirical study, we use $C = 10$, $ngram\_range = (1, 2)$ and $weighting = \text{TFIDF}$.

We report the performance of the BERT-based and LR-based STL runs and compare them with our BERT-based MTL run in Table 3. It shows that the BERT-based runs perform better than the LR-runs except for the marginal decrease in Cacc score. More importantly, our MTL run achieves substantial improvement in NDCG and AW scores over the BERT-based STL run. For example, the BERT_base + MTL run achieves the best scores in NDCG, AW-HC, AW-A and CF1-H. Although it can be seen that the BERT-based STL runs perform the best in prioritisation, this is at the cost of a loss of NDCG and AW performance. As a whole, our MTL scenario gains an advantage over the STL scenario not only in the overall effectiveness (as illustrated by the HarM score in the last column) but also in avoiding the need to train separate models for separate tasks.

---
[9]Unless stated otherwise, all pre-trained checkpoints mentioned in this paper refer to the official uncased ones provided by the transformers library (Wolf et al. 2019).

|  | Priority estimation | | | | | Info. type classification | | | |
| --- | --- | --- | --- | --- | --- | --- | --- | --- | --- |
|  | NDCG | AW-H | AW-A | PErr-H | PErr-A | CF1-H | CF1-A | Cacc | HarM |
| LR + STL | 0.4495 | -0.4856 | -0.2627 | 0.1718 | 0.2216 | 0.0898 | 0.1527 | **0.9113** | 0.2109 |
| BERT_base + STL | 0.4393 | -0.4057 | -0.2148 | **0.2402★** | **0.2758★** | 0.1084 | **0.1801** | 0.8960 | 0.2510 |
| BERT_base + MTL | **0.5101** | **-0.2689★** | **-0.1569★** | 0.1923 | 0.2544 | **0.1382★** | 0.1638 | 0.8937 | **0.2609** |

**Table 3.** Comparison between single task learning (STL) with Logistic Regression (LR) and BERT and our multi-task learning (MTL). The numbers in bold represent the highest performance in each column and those annotated with ★ indicates that the highest is "confident" compared to the next-highest in its column (Wilson Score Interval (Wilson 1927), $p < 0.05$). We describe the difference between $c_1$ and $c_2$ to be "confident" if their confidence intervals do not overlap.

### RQ2: Transformer selection and ensemble

In the pipeline of our approach, one important component is the transformer encoder selection. In the "Method" section above, we formulate our approach with the well-known transformer encoder BERT. However, other mainstream transformer encoders achieve state-of-the-art performance on various language understanding benchmarks (e.g., GLUE (A. Wang et al. 2018)) in the literature (Lan et al. 2020; Sanh et al. 2019; Clark et al. 2020). Since the introduction of BERT, the literature has seen many BERT variants being developed. Variants like DistilBERT, ALBERT, and ELECTRA have been developed to optimise various aspects of BERT such as memory consumption, computation cost or pre-training representation learning. Studies have shown their promising performance through fine-tuning in downstream tasks such as text classification and reading comprehension. To examine their capabilities for our problem, apart from the original BERT we fine-tune the following three transformer encoders on the TL dataset in our MTL approach and report their performance using the IS track metrics.

- DistilBERT (Sanh et al. 2019) is a distilled version of BERT. Compared to the original BERT, it has fewer trainable parameters and is thus lighter, cheaper and faster during training and inference. Given the size of the reduced model, the original paper reports that it still keeps comparative language understanding capabilities and performance on downstream tasks. We use its `base-uncased` pre-trained weights in our experiment.

- ALBERT (Lan et al. 2020) is a derivative of BERT that is mainly optimised for memory efficiency with two parameter-reduction techniques. First, it splits the embedding matrix in a BERT-like architecture into smaller matrices (separation of the embedding size and the hidden size), leading to reduced memory use while mathematically maintaining equivalent effect. In addition, it uses repeated layers, where the parameters are shared across different BERT hidden layers. This optimisation results in a smaller memory footprint although the computational cost remains similar to the original BERT (the same iteration through all hidden layers is still required). In our experiment, we use its official `base_v2` pre-trained checkpoint.

- ELECTRA (Clark et al. 2020) maintains essentially the same architecture and size as the original BERT except for a change in the embedding matrix as in ALBERT. What makes it stand out is that it adopts a different pre-training approach. Unlike the Masked Language Modeling (MLM) pre-training objective used in BERT, it trains a generator using the MLM objective to replace tokens in a sequence and meanwhile it trains a discriminator with the objective of identifying which tokens were replaced by the generator in the sequence. It is shown to outperform other transformers on language understanding benchmarks when using the same amount of computational power. In our experiment, we use its `base-discriminator` checkpoint.

The upper block of Table 4 presents the results of these encoders' performance[10]. As this shows, the BERT run has the same model size as the ELECTRA. However, there is no significant difference in performance between the runs when evaluated using the HarM score. Although BERT outperforms ELECTRA in AW, it loses this advantage in prioritisation. In comparison, DistilBERT and ALBERT are relatively smaller in size than BERT and ELECTRA. They still perform well overall, only slightly worse than the BERT and ELECTRA runs, which illustrates their effectiveness with much reduced model sizes (ALBERT in particular).

Of the individual runs, each scores the highest performance in at least one metric, with none showing a significant overall performance improvement over the others. In order to bring the power of these individual runs to our problem we propose a simple ensemble approach that combines the individual runs to jointly make predictions for information types and priority. We hypothesise that such an ensemble may leverage the distinct benefits of these diverse transformer encoders to achieve greater overall performance across both tasks.

---

[10]Note that the first row of Table 4 is the same as the last row of Table 3.

|  | Priority estimation |  |  |  | Info. type classification |  |  |  |
|---|---|---|---|---|---|---|---|---|
| Run variants | **NDCG** | **AW-HC** | **AW-A** | **PErr-H** | **PErr-A** | **CF1-H** | **CF1-A** | **Cacc** | **HarM** |
| *Individual transformer encoders* | | | | | | | | | |
| 1. BERT_base (110M) | 0.5101 | -0.2689 | -0.1569 | 0.1923 | 0.2544 | 0.1382 | 0.1638 | 0.8937 | 0.2609 |
| 2. DistilBERT_base (66M) | 0.4808 | -0.4533 | -0.2382 | 0.9004 | 0.1191 | 0.1376 | 0.1830 | 0.2110 | 0.2264 |
| 3. ELECTRA_base (110M) | 0.5042 | -0.4011 | -0.2122 | 0.2059 | 0.2801 | 0.1514 | 0.1742 | 0.8958 | 0.2689 |
| 4. ALBERT_base_v2 (11M) | 0.4669 | -0.4118 | -0.2190 | 0.1900 | 0.2720 | 0.0568 | 0.1707 | **0.9087** | 0.1923 |
| *Ensemble runs* | | | | | | | | | |
| EnsembleA (1+3) | **0.5207** | -0.2274 | -0.1406 | 0.1999 | 0.2560 | 0.1738 | 0.1796 | 0.8722 | 0.2836 |
| EnsembleB (2+4) | 0.4848 | -0.3212 | -0.1823 | 0.2081 | 0.2728 | 0.1407 | 0.2041 | 0.8844 | 0.2752 |
| EnsembleC (1+2+3) | 0.5206 | -0.1982 | -0.1282 | 0.2023 | 0.2589 | **0.1819** | 0.1909 | 0.8621 | 0.2919 |
| EnsembleD (1+2+3+4) | 0.5176 | **-0.1613**★ | -0.1148 | 0.2594★ | 0.2966 | 0.1754 | **0.2084** | 0.8545 | **0.3141**★ |

**Table 4.** Individual runs with different transformer encoders and ensemble runs that leverage the individual runs jointly making predictions for priority and information types. The number appended to the individual runs refers to the trained model size and the number appended to the ensemble runs indicates the combination of corresponding individual run indices. The numbers in bold represent the highest in each column and ★ indicates that it is "confident" compared to the next-highest in its column.

**Our ensemble approach**: Given a set of individual multi-task learners, $\{l_1, l_2, ..l_n\}$, the final priority prediction for a tweet is made from the priority predictions by $\{l_1, l_2, ..l_n\}$ according to a priority decision strategy, denoted by $P_{ds}$. We evaluated three options for $P_{ds}$, where $P_{ds} \in \{Highest, Average, Lowest\}$. *Highest* refers to always selecting the highest priority level given by any of the individual predictors. *Lowest* represents the opposite strategy. The *Average* scenario means taking an average score over all priority predictions in the union and then the final priority prediction is assigned based on this average score. The conversion between priority numeric score and level is applied via the mapping function, namely $m(\cdot)$ as introduced in Equation 11. Regarding information types, the final prediction for each tweet is made from the information type predictions by $\{l_1, l_2, ..l_n\}$ according to a information type decision strategy, denoted by $I_{ds}$. We evaluated two options for $I_{ds}$, where $I_{ds} \in \{Union, Intersection\}$. *Union* and *Intersection* refers to always selecting the union and intersection respectively of information types by the individual predictors.

|  | Priority estimation |  |  |  | Info. type classification |  |  |  |
|---|---|---|---|---|---|---|---|---|
|  | **NDCG** | **AW-H** | **AW-A** | **PErr-H** | **PErr-A** | **CF1-H** | **CF1-A** | **Cacc** | **HarM** |
| Union-Highest | 0.5170 | -0.1613 | -0.1148 | 0.2594 | 0.2966 | 0.1754 | 0.2084 | 0.8545 | 0.3140 |
| Union-Average | 0.5066 | -0.2489 | -0.1491 | 0.2475 | 0.274 | 0.1754 | 0.2084 | 0.8545 | 0.3036 |
| Union-Lowest | 0.4896 | -0.5824 | -0.2932 | 0.1302 | 0.2102 | 0.1754 | 0.2084 | 0.8545 | 0.2369 |
| Intersection-Highest | 0.5178 | -0.1613 | -0.1148 | 0.2342 | 0.2715 | 0.0303 | 0.1105 | 0.9291 | 0.1387 |
| Intersection-Average | 0.5061 | -0.2489 | -0.1491 | 0.2184 | 0.2485 | 0.0303 | 0.1105 | 0.9291 | 0.1362 |
| Intersection-Lowest | 0.4888 | -0.5824 | -0.2932 | 0.1321 | 0.2215 | 0.0303 | 0.1105 | 0.9291 | 0.1233 |

**Table 5.** Evaluation results of our `EnsembleD` run with varying strategies for merging information types and priority levels. Each row is named as $x - y$ where $x$ is the information type strategy, i.e. $I_{ds}$ and $y$ is the priority level strategy, i.e. $P_{ds}$.

In choosing $P_{ds}$ and $I_{ds}$, we initially conducted experiments using an ensemble of all four individual runs from the top of Table 4 and the results are reported in Table 5. The results show that the *Union* runs substantially outperform the *Intersection* runs in information feed categorisation while yielding the same scores in the remaining metrics. For $P_{ds}$, we see an increased performance in ranking, alert worth, and prioritisation as it changes from *lowest* to *highest*. Based on the results, in our subsequent experiments, we set $I_{ds}$ to be *Union* and $P_{ds}$ to be *Highest* as this combination gives the best performance across the metrics.

With this setup, next we experiment using different sets of $\{l_1, l_2, ..l_n\}$ and the lower block of Table 4 demonstrates their performance. The EnsembleA runs combining the two relatively large models, BERT and ELECTRA, while EnsembleB combines DistilBERT and ALBERT. We see that each ensemble has overall performance superior to its component models. This indicates that the ensemble approach adds benefit to the performance by leveraging the predictions of individual runs. The best-performing run among all experimental runs so far reported in our study is EnsembleD, with the HarM score reaching 0.3141 as well as achieving strong scores in individual metrics except for in Cacc. However, we note that Cacc is arguably the least important metric due to the heavy imbalance in the TREC-IS information types and the usual problems with accuracy-based metrics in such a scenario. A naïve run that predicts all tweets to have all information types will achieve a Cacc score of approximately 0.94, while being useless in practical terms and performing extremely poorly in the other metrics. To examine the performance of the ensemble runs in separate tasks, we found that the ensemble runs outperform the single model based runs in both priority estimation and information type classification tasks. Although the overall performance increases as more individual models combined, our experiments found that the increase becomes marginal when there are more models combined than the EnsembleD. Hence, we take EnsembleD as the best run of our system considering its effectiveness and size.

### RQ3: TREC-IS track participation

We have reported that our MTL-based runs have an advantage over both the BERT-based and LR-based STL baselines. In addition, we have proposed a simple ensemble technique that leverages the individual MTL runs to jointly make predictions for information types and priority levels. This ensemble technique further improves performance as compared to the individual runs. The remaining research question relates to the performance of such an approach compared to the state of the art. This is measured by comparing it to participating runs in the most recent 2020-A edition of the TREC-IS track. This edition proposed two tasks, task 1 and task 2. The only difference between task 1[11] and task 2 is that task 2 uses a reduced set of 12 information type labels, which includes 11 important information types[12] from the 25 used in task 1 (see Table 1) with the remaining 14 combined into the single category "Other-Any". Thus task 2 emphasises a run's performance in identifying the information types that are most closely related to emergency response. Due to the common features shared between the two tasks, any runs submitted to task 1 are also evaluated in task 2.

In the 2020-A edition TREC-IS, we submitted several runs, of which `Our_Run1` is our MTL-based run that is similar to the `BERT_base+MTL` run[13]. Figure 3 plots the returned results of `Our_Run1` and the top participating runs, which are evaluated for both task 1 and 2. We also include our `EnsembleD` to show how it performs compared to the submitted runs. Although `EnsembleD` was not officially submitted, it was subsequently evaluated using IS's official evaluation script. The plotted results present the performance of the top participating runs from the four aspects: Ranking, Alert Worth, Information Feed Categorisation and Prioritisation.

When examining the participating runs, it appears that they frequently achieve high scores in some metrics at the cost of lower scores in others. For example, the `elmo` runs relatively outperform `Our_Run1` in Alert Worth (Figure 3c and 3d) but fall far behind in Ranking and Information Feed Categorisation. In contrast, the `sub` runs achieve good scores that are near to `Our_Run1` in Information Feed Categorisation (Figure 3e and 3f) but not in the Alert Worth metrics. Despite the loss in Alert Worth to the `elmo` runs, `Our_run1` outperforms the top participating runs in the rest of metrics for both task 1 and task 2 with the only exception of a marginal loss to one `elmo` run in Task 1 Prioritisation (Figure 3g). `Our_run1` also achieves the highest HarM score, which implies overall best performance. However, there remains a strong argument that in practical terms different submitted runs are preferable in different situations, depending on the needs of the emergency responders, and that no overall best-performing system has been satisfactorily identified.

The `EnsembleD` run, however, achieves state-of-the-art performance in almost every metric, substantially outperforming the participating runs in most cases. Figure 3 indicates that there are only two evaluation figures (other than the less important Cacc discussed previously) where `EnsembleD` does not have the highest performance, with the difference being minor in both cases (the CF1-A from Figure 3e and PErr-H from 3g). Given the tendency of participating runs to achieve imbalanced performance across the individual metrics, our `EnsembleD` stands out as a good choice with regards to its effectiveness in different aspects of emergency response, across the range of metrics. Hence, we consider `EnsembleD` to be a strong baseline for further work into this problem by the community in the future.

### ERROR ANALYSIS

Although our approach-based runs (EnsembleD in particular) achieve effective performance in multiple aspects, we are interested in knowing what types of errors are manifested in our system. We subsequently conduct a qualitative error analysis for our system, aiming to provide insights of solving the challenges in this field for the community. Here, using our best-performed run, (i.e., EnsembleD), we focus on the information type categorisation task due to its importance in emergency response. Table 6 presents some interesting examples of wrong information types predictions by EnsembleD. "IT prediction" is the set of information types assigned by EnsembleD, whereas "IT ground truth" lists the information types assigned by human assessors as part of the IS track.

---

[11]The results we have reported so far are from Task 1.
[12]The information types include the 6 actionable information types in Table 1 plus Request-InformationWanted, CallToAction-Volunteer, Report-FirstPartyObservation, Report-Location and Report-MultimediaShare.
[13]The difference is that `Our_Run1` is trained on the entire training set without leaving out the development set that we use in `BERT_base+MTL`.

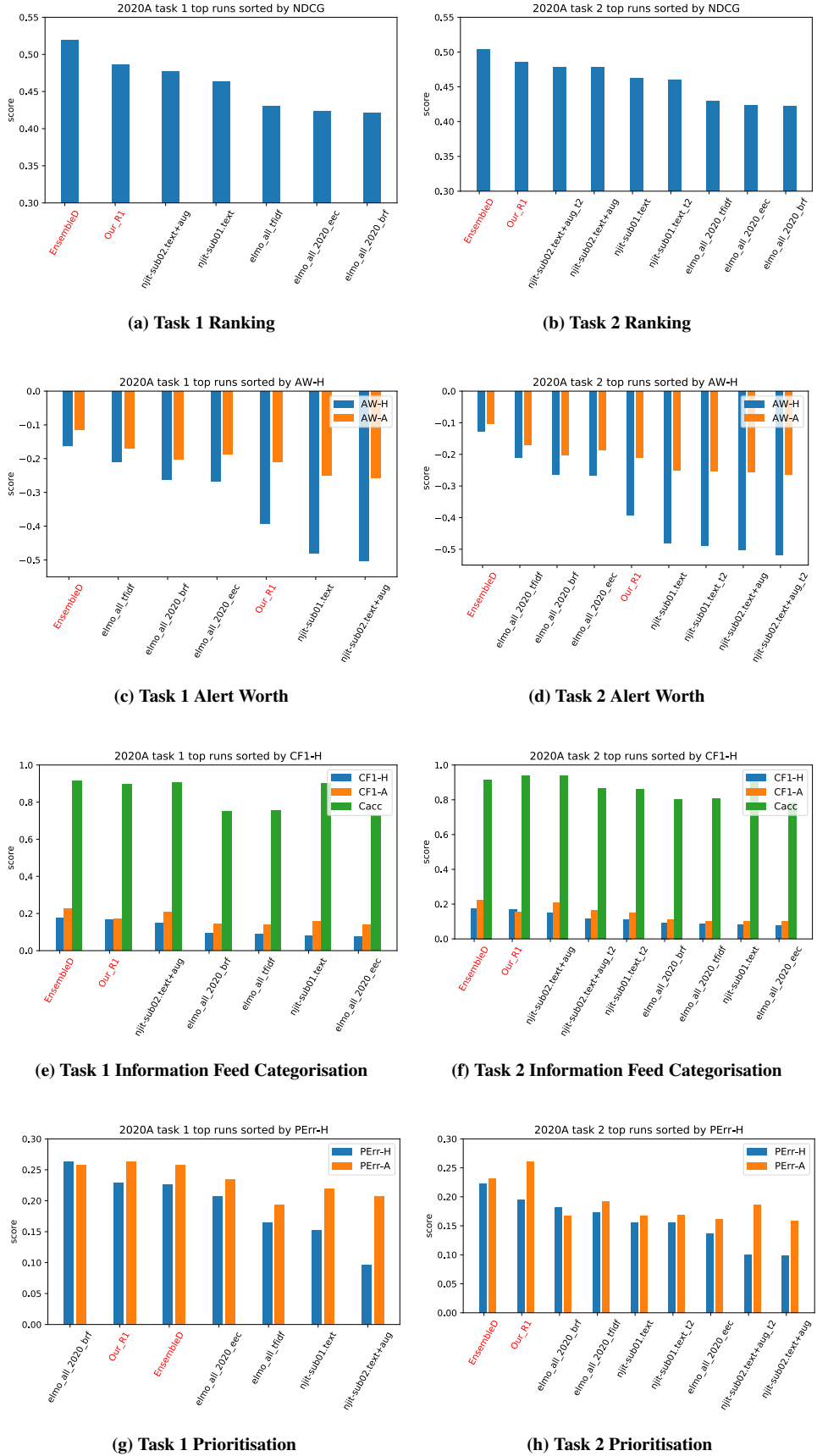

Figure 3. Performance comparison between TREC-IS 2020A top participating runs for both task 1 and 2, which are evaluated from four major aspects: Ranking, Alert Worth, Information Feed Categorisation and Prioritisation. Our runs are annotated in red and the rest are other participating runs.

| Id | Event | Text | IT prediction | IT ground truth |
|---|---|---|---|---|
| #1 | gilroygarlicShooting2020 | 6-year-old killed in Gilroy Garlic Festival Shooting https://t.co/MYgvoneYdC | Report-Factoid, Report-Location, Report-MultimediaShare, Report-News, Report-ThirdPartyObservation | Report-Factoid, Report-Location, Report-MultimediaShare, Report-News |
| #2 | gilroygarlicShooting2020 | Volunteers from Muslim-faith based charity @pennyappeal have turned up at Chapel school with loads of supplies for people evacuated from #WhaleyBridge. Good on ya, lads. https://t.co/eBMM3evPAL | CallToAction-Donations, Report-Hashtags, Report-Location, Report-MultimediaShare, Report-ServiceAvailable | Report-Hashtags, Report-Location, Report-MultimediaShare, Report-ServiceAvailable, Report-ThirdPartyObservation |
| #3 | hurricaneBarry2020 | Rain water at Mississippi and Santa fe #Denver @9NEWS https://t.co/a8eekFIODx | Report-Hashtags, Report-Location, Report-MultimediaShare, Report-News | Other-Irrelevant |
| #4 | baltimoreFlashFlood2020 | I'm at Bronycon 2019 in Baltimore, MD https://t.co/oVXxmZ4JID | Other-Irrelevant | Request-SearchAndRescue |

**Table 6. Examples of error analysis for information types**

The first two examples are interesting in how the "Report-ThirdPartyObservation" IT is handled. In the first example, EnsembleD chose this IT but it was missing from the ground truth ITs. In contrast, for the second example EnsembleD did not choose this IT but the ground truth included it. It is arguable that all tweets containing information about a crisis ought to be tagged as first-party or third-party observations. That said, is interesting that the image associated with #2 suggests that this may actually be a first-party observation with the author of the tweet posting an image of the volunteers that are mentioned. Additionally it is interesting to note that EnsembleD chose the "CallToAction-Donations" IT. Although this tweet is not explicitly a call to action, the referenced Twitter account (@pennyappeal) represents a charity appeal for crisis situations. Both of these observations indicate that context and subtlety make IT classification a difficult task.

Geographical context is important in example #3. The human assessor has marked it irrelevant as it relates to Denver, Colorado whereas the major effects of Hurricane Barry were felt primarily in Louisiana. The attached image shows flood water within the same time period as the hurricane in question. As such, had it been within the geographical area in question, many of the ITs chosen by EnsembleD would likely be correct. This example also further illustrates another challenge with regard to geographical context. Without context, "Mississippi" could refer to a US state or a river and "Santa Fe" is the capital city of the US state of New Mexico. However, in this instance these names refer to street names in the city of Denver, Colorado, where the picture of floodwater was taken. Although this was not the reason why this particular tweet was misclassified it indicates a further challenge, suggesting that language models alone will not be the sole solution to this problem and that the incorporation of knowledge maps and other ontologies may be necessary to lend necessary context to IT classification systems.

The final example seems to simply represent human error in the process of assigning ITs. This tweet refers to the user attending a conference that ended two days prior to the flash flood that is being referred to[14]. This example serves to emphasise the limitations inherent in using human assessors for comparison. Personal opinion, outside context and straightforward errors mean that it is important for systems to be examined qualitatively as well as reporting evaluation metrics.

## CONCLUSION AND FUTURE WORK

In this paper, we present a transformer-based multi-task learning approach for crisis tweet classification and prioritisation, which is a crucial problem in emergency response. We empirically present our approach's leading performance as compared to single task baselines as well as competitive participating runs from the IS track. Additionally, we introduce a simple ensemble approach that leverages multiple multi-task learners for categorising the crisis-related test tweets. This ensemble approach turns out to be more effective than its constituents, and achieves state-of-the-art performance when compared with participating runs in the TREC-IS track.

---

[14] It is possible that the assessor felt that anybody finding themselves attending Bronycon is necessarily in need of rescue, but that question is outside the scope of this paper.

Regarding the effectiveness, there is still much room to improve. We offer our approach as a baseline for future work on this problem. Our approach is currently limited to using the raw text of tweets. Per McCreadie et al. 2020, linked content such as the web pages, or images posted along with the tweets are likely to improve performance. Hence, future work will incorporate this linked information into our current approach. In addition, since our current approach only focuses on two tasks from a single corpus, we plan to extend our approach to support more tasks from multiple corpora. Although effectiveness is an essential performance measurement for an emergency response system, another aspect to which the same importance should be given is efficiency. When employing an emergency response system in a real-world situation, the speed of message handling is crucial because social media message quantities are usually enormous during a crisis. Our approach has not yet been examined regarding its efficiency, which also requires further exploration.

## ACKNOWLEDGEMENTS

This project has received funding from Enterprise Ireland and from the European Union's Horizon 2020 research and innovation programme under the Marie Skłodowska-Curie grant Agreement No 713654